# Fast Algorithm of High-resolution Microwave Imaging Using the Non-parametric Generalized Reflectivity Model

Long Gang Wang; Lianlin Li, *Senior Member*; Tie Jun Cui, *Fellow*

Abstract:

This paper presents an efficient algorithm of high-resolution microwave imaging based on the concept of generalized reflectivity. The contribution made in this paper is two-fold. We introduce the concept of non-parametric generalized reflectivity (GR, for short) as a function of operational frequencies and view angles, etc. The GR extends the conventional Born-based imaging model, i.e., single-scattering model, into that accounting for more realistic interaction between the electromagnetic wavefield and imaged scene. Afterwards, the GR-based microwave imaging is formulated in the convex of sparsity-regularized optimization. Typically, the sparsity-regularized optimization requires the implementation of iterative strategy, which is computationally expensive, especially for large-scale problems. To break this bottleneck, we convert the *imaging problem* into the problem of *physics-driven image processing* by introducing a dual transformation. Moreover, this image processing is performed over overlapping patches, which can be efficiently solved in the parallel or distributed manner. In this way, the proposed high-resolution imaging methodology could be applicable to large-scale microwave imaging problems. Selected simulation results are provided to demonstrate the state-of-art performance of proposed methodology.

# 1. Introduction

Microwave imaging is a very promising tool of nondestructive examination in various civil and military applications, such as, geosciences, medicine and other areas [1-6]. In principle, it is known as the electromagnetic inverse scattering problem, which aims at retrieving the distribution of electrical parameters (reflectivity in this paper) of probed objects from its associated scattering fields by solving a nonlinear optimization problem [1]. Over past decades, some optimization schemes have been developed and intensively investigated, for instance, Newton-Kantorovich method [7], distorted Born iterative method [8], Gauss-Newton inversion method [9], contrast inverse method [10] and modified gradient method [11], etc. A detailed comparison of these methods can be found in [12] and references herein. However, it is a consensus that the electromagnetic inverse scattering problem is computationally expensive even for moderate-scale problem, and thus has found limited practical success only at very low frequencies, as argued in [13]. Therefore, one practical strategy is to pursue an approximate solution to the rigorous electromagnetic inverse scattering problem, where the trade-off between the solution accuracy and efficiency are needed according to practical requirements.

Roughly speaking, there are two approximate strategies explored widely in practical applications [e.g., 13,14, 15, 35, 36]: migration and inversion. Migration relies on so-called *exploding* source model, assuming that the probed object consists of a collection of independent point-like targets. The migration produces an image by performing the (filtered) back-propagation. Many efficient migration algorithms have been developed by now, for instance, the Range-Doppler, Chirp-Scaling, Omega-K, and so on [14, 15]. The migration assumes that the data is linearly related to the reflectivity of object by the Fourier transform, which is approximately valid for the far-field imaging with very narrow frequency band and viewing-angle scope. Technically, migration can be realized by performing fast (inverse) Fourier technique, and thus it has very low computation complexity at the cost of scarifying the image quality. Inversion is based on well-known Born approximation [1, 2, and 8, 35, 36], i.e., single-scattering approximation, implying that the undergoing (extended) object should be very weak. Here, the weak object implies $k_0 \Delta \varepsilon_r D \ll 1$ [1], where $k_0$ denotes the operational wavenumber in the background medium, $\Delta \varepsilon_r$ is the difference of relative permittivity between the probed object and background medium, and *D* is the diameter of smallest sphere encircling probed object. The inversion produces an image by solving a set of linear equations, which has heavier computational cost than migration. However, the distinct advantage of inversion over migration is that it is capable of producing high-resolution, even super-resolution images for weak objects [e.g., 16, 17, 35, 36].

A common drawback of both migration and inversion is that the image resolution is relatively low for general objects due to the use of oversimplified imaging models mentioned above, leading to the occurrence of ghost images, and thus causing very heavy burden on post-process of object identification and classification [18, 19]. To overcome this shortcoming, the imaging problem was treated in the context of regularized optimization, where some prior knowledge on probed objects can be exploited in a flexible way. Since the information contained in the output reconstruction is composite of prior knowledge and data, it could, in principle, produce the image with enhanced resolution, and could be used directly in enhanced classification and recognition without an intermediate reconstruction. One representative example is the super-resolution technique developed by Luttrell et al in the community of radar imaging [20]. More interestingly, with the development of compressive sensing (CS) theory, the regularized imaging technique has been attracting researchers' attentions. In literatures, such type of resulting methodology is also referred to as the CS-based imaging or sparsity-promoted imaging. For example, Cetin et al developed a feature-enhanced imaging technique by which the profile of probed object edge could be enhanced by imposing the total-variation (TV) regularization on the probed object with piecewise property [21]. Zhu and Bamler et al studied the application of sparsity-regularized imaging approach in the tomographic SAR problems [22]. We would like to refer to [23] for a comprehensive overview on this topic.

Although the sparsity-regularized imaging technique has been investigated intensively in the area of electromagnetic imaging over the past decade, some important issues remain open. Most of them rely on the same signal model as those adopted by migration and inversion, where the realistic interaction between the wavefield and object cannot be well accommodated. Therefore, the produced image could be seriously distorted by artifacts. To tackle this problem, we extend the Born model limited to weak objects into that handling general objects by introducing the concept of *generalized reflectivity (GR)*. It should be emphasized that the GR can be regarded as the equivalent current induced inside object normalized by the incident wavefield. In other words, the GR behaviors as a function of frequencies and view angles. Apparently, the dimension of the GR to be reconstructed will be remarkably enlarged when considering various sampling frequencies and/or view angles. To tackle this issue, the sub-array technique [23] and the mixed norm [27] are exploited, which is inspired from two fundamental observations investigated in section 2. Afterward, we formulate microwave imaging into the one that consists of retrieving the GR by solving a sparsity-regularized optimization problem. The sparsity-regularized imaging problem has no closed-form solution and requires the implementation of iterative strategy, which is computationally intensive, especially for large-scale problems. To break this bottleneck, we convert the *imaging problem* into the problem of *physics-driven image processing* by introducing a dual transform. Moreover, the image processing is performed over overlapping patches in the parallel or distributed manner. In this way, the imaging speed will be remarkably accelerated while maintaining acceptable reconstruction resolution.

The rest of this paper is organized as follows: Section 2 describes the principle of proposed generalized reflectivity model. Section 3 investigates two sparsity-driven imaging algorithms: the first-order iteration algorithm and the fast imaging algorithm for relieving heavy computational cost. In section 4, selected numerical simulations are presented to demonstrate the state-of-art performance of proposed methodology. Finally, section 5 summarizes this paper.

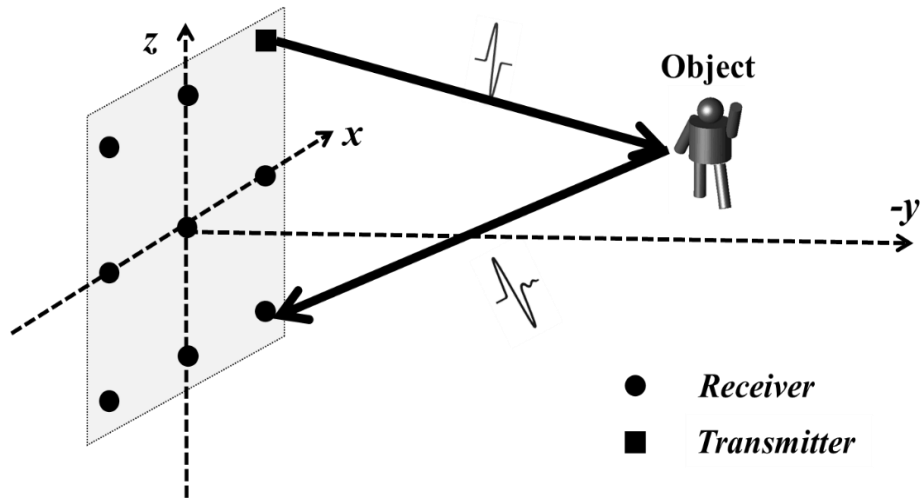

Fig.1 The configuration of MIMO microwave imaging system with T transmitters (square) and S receivers (dots).

## II. The GR-based Imaging Model

This section elaborates on the concept of generalized reflectivity model used in this paper. For the purpose of illustrating the principle of proposed methodology, we consider a popular configuration of microwave imaging, i.e., the multiple-input and multiple-output (MIMO), as sketched in Fig. 1, where T transmitters distributed in the plane y=0 work sequentially, and $S$ receivers are allocated in the plane y=0 to collect scattered fields emerged from the investigated object as well. Here, we would like to say that the developed methodology and conclusions are applicable to other imaging scenarios.

When the *t*th transmitter at $r_t$ (*t=1,…,T*, *T* is the total number of transmitters) sends out a frequency-dependent illumination signal $\tilde{s}(\omega)$ towards the object, a current denoted by $J(r;\omega,r_t)$ will be induced inside the object, where $r \in V$, *V* is the investigation domain and $\omega$ is the working angular frequency. Here, the arguments of $\omega$ and $r_t$ highlight the dependence of currents on operational frequencies and view angles, respectively. In terms of the electrical integral equation, the electrical field radiated from the induced current $J(r;\omega,r_t)$ reads [25]:

$$\mathrm{E}(r_s;\omega,r_t) = i\omega\mu_0 \int_V \mathrm{d}r' \, \mathbf{G}(r_s,r';\omega) \cdot J(r',\omega;r_t) \tag{1}$$

where $r_s$ indicates the location of the *s*th receiver ($s = 1,2,…,S$, $S$ is the total number of receivers), and $\mathbf{G}(r,r';\omega)$ is the dyadic Green's function in free space [25], i.e.,

$$\mathbf{G}(r,r';\omega) = \left(\mathbf{I} + \frac{1}{k_0^2}\nabla\nabla\right) g(r,r';\omega)$$

$$= \frac{e^{ik_0 R}}{4\pi R}\left[\left(3\frac{\mathbf{R}\otimes\mathbf{R}}{R^2} - \mathbf{I}\right)\left(\frac{1}{k_0^2 R^2} - \frac{i}{k_0 R}\right) + \left(\mathbf{I} - \frac{\mathbf{R}\otimes\mathbf{R}}{R^2}\right)\right] \tag{2}$$

where $R = |r - r'|$, $g(r,r';\omega)$ is the scalar three-dimensional Green's function in free space, $k_0 = \omega/c$ denotes operational wavenumber. Introduce a $3 \times 3$ tensor $\bar{\bar{\mathcal{R}}}(r';\omega;r_t)$ such that $J(r';\omega;r_t) = \bar{\bar{\mathcal{R}}}(r';\omega;r_t) \cdot E_{in}(r';\omega;r_t)$, where $E_{in}$ indicates the incident wavefield in the absence of objects. Now, Eq. (1) can be rewritten as:

$$E(r_s;\omega,r_t) = i\omega\mu_0 \int_V \mathrm{d}r' \, \mathbf{G}(r,r';\omega) \cdot \bar{\bar{\mathcal{R}}}(r';\omega;r_t) E_{in}(r';\omega;r_t) \tag{3}$$

Note that $\bar{\bar{\mathcal{R}}}(r';\omega;r_t)$ is exactly the conventional reflectivity when it is assumed to be free of operational frequency and viewing angle, i.e., $\bar{\bar{\mathcal{R}}}(r';\omega;r_t) = \mathcal{R}(r')\bar{\mathbf{I}}$, where $\bar{\mathbf{I}}$ is the unit dyadic. Consequently, Eq. (3) reduces into standard Born model. For this reason, we would like to refer to $\bar{\bar{\mathcal{R}}}(r';\omega;r_t)$ as the GR, which is a function of operational frequencies and view angles. Although the GR defined through $J(r';\omega;r_t) = \bar{\bar{\mathcal{R}}}(r';\omega;r_t) \cdot E_{in}(r';\omega;r_t)$ has formal similarity with well-known quasi-linear approximation [26], they are completely different in determining $\bar{\bar{\mathcal{R}}}(r';\omega;r_t)$. For the quasi-linear model, $\bar{\bar{\mathcal{R}}}(r';\omega;r_t)$ is assumed to be free of operational frequencies and view angles, i.e., $\bar{\bar{\mathcal{R}}}_{QL}(r';\omega;r_t) = \bar{\bar{\mathcal{R}}}_{QL}(r')$, and it is determined by jointly solving the so-called *data* equation and *state* equation in literatures of inverse scattering [26]. Here, the data equation corresponds to Eq. (3), while the state equation corresponds to Eq. (3) but the left hand being $E(r;\omega,r_t) - E_{in}(r;\omega,r_t)$ subject to $r \in V$. On the contrary, the $\bar{\bar{\mathcal{R}}}(r';\omega;r_t)$ for the GR model behaviors as a function of operational frequencies and view angles. Furthermore, the GR-based approach aims at retrieving the GR by solving the data equation, Eq. (3), in the context of sparsity-regularized optimization problem. Note that the prior knowledge on $\bar{\bar{\mathcal{R}}}(r';\omega;r_t)$, such as the strong coherence and joint sparsity of $\bar{\bar{\mathcal{R}}}(r';\omega;r_t)$ over different frequencies and view angles, has been incorporated into the reconstruction algorithm.

In numerical implementation, the whole investigation domain *V* is divided into *N* voxels, and both $E_{in}(r';\omega;r_t)$ and $\bar{\bar{\mathcal{R}}}(r';\omega;r_t)$ are constant within each voxel. Additionally, the number of discrete frequencies is assumed to be *F*. Now, assuming that $y_{f,t} = E(r_s;\omega,r_t)$ is stacked as an *S*-length column vector of a subset of acquired data for a given operational frequency indexed by *f* and a transmitter indexed by *t*. Similarly, $\mathbf{A}_{f,t} = i\omega\mu_0 \mathbf{G}(r,r';\omega) E_{in}(r';\omega;r_t)$ corresponds to the measurement matrix with size of *S* by *N*, and $x_{f,t} = \bar{\bar{\mathcal{R}}}(r';\omega;r_t)$ is stacked as an *N*-length column vector of generalized reflectivity. Then Eq. (3) can be reformulated in a compact form:

$$y_{f,t} = \mathbf{A}_{f,t} x_{f,t} + n_{f,t} \tag{4}$$

$$f = 1,2,\ldots,F, \text{ and } t = 1,2,\ldots,T$$

Herein, $n_{f,t}$ is an *S*-length column vector accounting for noise, measurement error, and other possible uncertainties.

Microwave imaging aims at retrieving $\{x_{f,t}\}$ from $\{y_{f,t}\}$ by solving Eq. (4). Compared to Born-approximation model, Eq. (4) needs to be solved for different generalized reflectivity associated with different operational frequencies and view angles. In this way, there are much more unknowns solved in Eq. (4) than that of Born-approximation model. This is a challenging problem since it is extremely ill-posed. To tackle this issue, we explore two fundamental observations:

(*a*)The equivalent currents induced inside object have strong coherence over different operational frequencies and/or view angles, especially for the case that the object is not very strong, which is related to the concept of low-rank matrix [27, 28]. Such low-rank prior on the generalized reflectivity can be enforced by exploiting the sub-array technique [27]. For discussion convenience, assuming that the whole operational frequency band and viewing aperture are uniformly divided into a series of overlapping sub-bands and sub-apertures [19, 27], which are indexed by $k$=1,2,…,$K$ in order, where $K$ is the total number of sub-apertures or/and sub-bands. Then, for the $k$th sub-band or/and sub-aperture, the induced currents are approximated to be completely coherent. So the notation $\mathbf{A}_{f,t}$, $y_{f,t}$ and $x_{f,t}$ in Eq.4 can be replaced by $\mathbf{A}_{(k)}$, $y_{(k)}$ and $x_{(k)}$ ($k$=1,2,…,$K$) respectively. Correspondingly, Eq. (4) becomes:

$$y_{(k)} = \mathbf{A}_{(k)} x_{(k)} + n_{(k)} \quad k\text{=1,2,..,}K \tag{5}$$

where the value of K will be significantly smaller than $F \times T$. In this way, the number of unknowns is considerably reduced, and thus the degree of ill-posedness of Eq.(4) will be relieved.

(*b*)It is noticed that the induced current $J(r'; \omega; r_t)$ is supported by the object regardless of operational frequencies and view angles, and thus the induced currents for different operational frequencies and view angles share the common support. Therefore, if the object is sparse, such property can be characterized by

so-called joint sparsity, imposed mathematically by the (p,q)-mixed norm [23] defined as:

$$\Omega(X) = ||X||_{P,q} = \left(\sum_{n=1}^{N}(\sum_{f,t}|x_{f,t}(n)|^p)^{q/p}\right)^{1/q}, \ 2 \leq p \leq \infty, 0 \leq q \leq 1 \quad (6)$$

where $X$ denotes the matrix stacked by vectors $\{x_{f,t},\}$. As a consequence, the inverse problem of Eq. (5) can be casted into following sparsity-regularized optimization, i.e.,

$$min_X[\varphi(X) + \gamma\Omega(X)] \quad (7)$$

Here, $X$ indicates a matrix stacked by $K$ column vectors $\{x_{(k)}, k = 1,2,...,K\}$. In Eq. (7), the first term $\varphi(X) = \frac{1}{2}\sum_{k=1}^{K}||y_{(k)} - A_{(k)}x_{(k)}||_2^2$ indicates data fidelity, the non-smoothness penalty term $\Omega(X)$ is sparsity term, and $\gamma$ is a regularized coefficient.

## III. The Reconstruction Algorithm

### III. A First-order Iterative Algorithm

Over past years, a great amount of algorithms have been developed, which can be used to tackle Eq. (7). In this paper, we would like to adopt popular first-order iterative methods to solve Eq. (7) due to following considerations [29-31]. First, as implied by its name, first-order methods only involve the operation of gradient, which has low computational cost. Second, first order methods support easy-implementation solution to not only smooth problem but also non-smooth (non-convex) problem. Usually, $\Omega(X)$ defined above is non-smooth. It has been demonstrated that the non-smooth optimization problem (7) can be solved nearly as efficiently as smooth problems, provided that the computational of the proximal operator defined below is tractable in a closed form. Finally, first order methods provide a flexible framework to distribute optimization tasks and perform computations in the parallelized or distributed manner.

First-order iterative methods rely on a fundamental fact that the smoothness term $\varphi(X)$ is L-Lipschitz smooth which implies that the inequality of $\varphi(X) \leq \varphi(X_0) + \langle \varphi(X_0), X - X_0 \rangle + \frac{L}{2}||X_0 - X||_F^2$ holds up for any $X_0, X \in dom(\varphi)$ [30, 31]. Then, first-order methods admit that Eq. (7) can be solved by starting at some initial point and performing following iterative equation [31]:

$$X^{(n+1)} = min_X \left[ \varphi(X^{(n)}) + \langle \nabla\varphi(X^{(n)}), X - X^{(n)} \rangle + \frac{L}{2}||X - X^{(n)}||_2^2 + \gamma\Omega(X) \right]$$

$$= min_X \left[ \frac{1}{2}||X - P(X^{(n)})||_2^2 + L^{-1}\gamma\Omega(X) \right] \quad (8)$$

$$\equiv Prox_{L^{-1}\gamma\Omega}(P(X^{(n)})) \quad (9)$$

where $Prox_{L^{-1}\gamma\Omega}$ is so-called proximal operator [31], $P(X^{(n)}) = X^{(n)} - L^{-1}\nabla\varphi(X^{(n)})$ is exactly the gradient-descent iterative solution of $\varphi(X)$ in the proximity of $X^{(n)}$. Interestingly, note that $P(X^{(n)})$ corresponds to the solution obtained by performing standard back-projection algorithm. However, the proximal operation corresponds to the denoise-like image processing, where the prior knowledge on probed objects has been taken into account through $\Omega(X)$. In language of microwave imaging, the iterative solution of Eq.(8) is the *closed-loop* consisting of the back-projection and image processing, as illustrated in Fig. 2, as opposed to conventional *open-loop* processing strategies in the area of radar imaging. In addition, the update step size (or Lipchitz constant *L*) can be easily determined in the closed-form by using the linear search strategy. Finally, we can achieve the whole imaging algorithm of solving Eq. (7), as summarized in Algorithm 1.

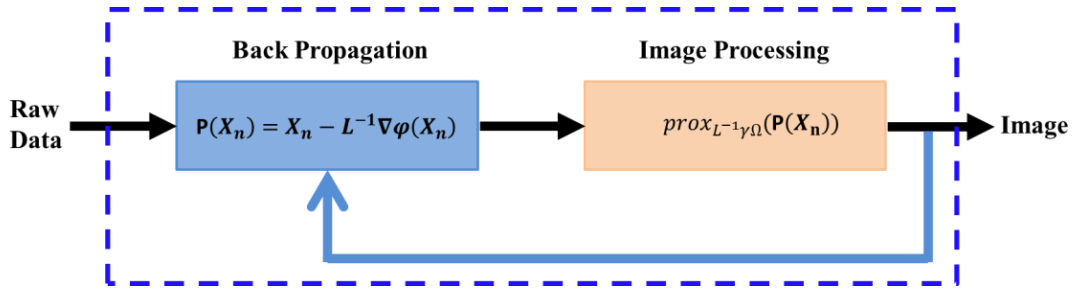

**Figure 2.** The closed-loop imaging strategy consisting of the back-projection and image

processing.

Algorithm 1: The procedure of first-order iterative algorithm for solving Eq. (7)
___

Input : $y_{(k)}$ $(k = 1,2 \cdots K)$ and $A_{(k)}$

Output : Reflectivity: $X$
___

- Initialize $x_{(k)}= 0$ and $tmp\_y_{(k)} = A_{(k)} \cdot x_{(k)}$
- Do while not achieving at some stop criterion:
  For k = 1:K
  ➢ Step 1: Compute $d_{(k)} = \nabla \varphi(x_{(k)})$
  ➢ Step 2: Compute the step factor $\alpha_{(k)} = \dfrac{-\|d_{(k)}\|_2^2}{\|A_{(k)} d_{(k)}\|_2^2}$
  ➢ Step 3: Update the *n*th $x_{(k)}$, and $x_{(k)}^n = x_{(k)}^{n-1} + \alpha_{(k)} d_{(k)}$
  End
  ➢ Step 4: Fuse $x_{(k)}$ into $X$ via Eq.(9)
  ➢ Step 5: Check the error
- End do while
___

## *III. B Fast Imaging Algorithm*

From the standpoint of computational efficiency, Algorithm 1 outlined above has a drawback that it call all measurements and involve all unknowns at each iteration. For this reason, it has low computational efficiency and is limited to cope with small-scale or moderate-scale problems. As a matter of fact, mainly due to this reason, the super-resolution technique developed in community of radar imaging gets very limited practical popularity. In order to break this bottleneck, we transform the imaging problem into the physics-driven image processing problem, and the operation of image processing is performed on the overlapping patches [32, 33]. As a consequence, the imaging problem is decomposed into a series of *parallel* small-scale image processing sub-problems [32, 33]. Therefore, this methodology can be applicable to very large-scale high-resolution electromagnetic imaging problem while maintaining high imaging quality. The details about it are discussed as follows.

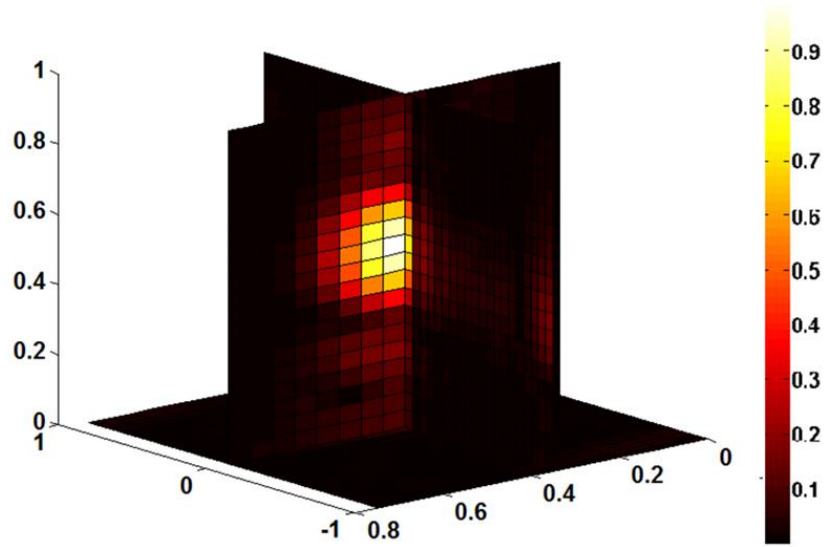

**Fig.3** The majority of energy of $P_{(1)}(j, 6728)$ (the fixed unit is the 6728th resolution unit and k=1) as a function of $r_j$ is concentrated around $r_{6728}$, where the $\mathbf{P}_{(k)}$ is coming from the following example 1.

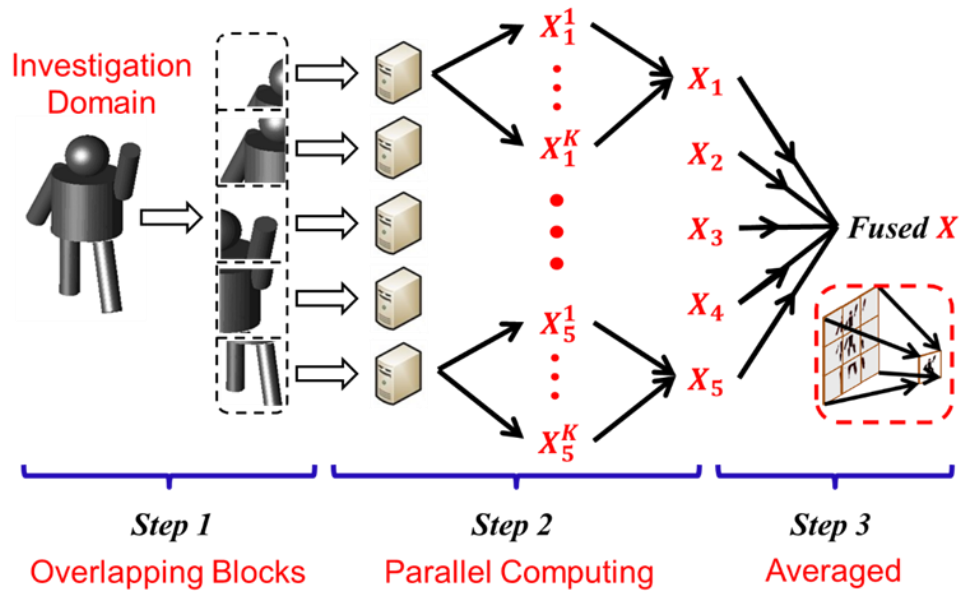

**Fig.4** The flow chart of fast imaging algorithm over overlapping patches.

For the kth reflectivity component $x_{(k)}$, recall its measurement equation as:

$$y_{(k)} = A_{(k)}x_{(k)} + n_{(k)} \qquad k = 1,2,...,K \tag{10}$$

Introducing a dual transform $T_{(k)}: dom(y_{(k)}) \rightarrow dom(x_{(k)})$, where *dom* indicates the domain of undergoing argument, Eq.(10) becomes:

$$T_{(k)}y_{(k)} = T_{(k)}A_{(k)}x_{(k)} + T_{(k)}n_{(k)} \qquad k = 1,2,...,K \tag{11}$$

A natural choice for $T_{(k)}$ is $(A_{(k)})^*$ or its approximation (e.g., far-field approximation), where the superscript * means the transpose conjugate. This paper uses $T_{(k)} = (A_{(k)})^*$ for simplicity. Introducing notations of $z_{(k)} = T_{(k)}y_{(k)}$, $P_{(k)} = T_{(k)}A_{(k)}$ and $\tilde{n}_{(k)} = T_{(k)}n_{(k)}$, then Eq. (12) can be rewritten as

$$z_{(k)} = P_{(k)}x_{(k)} + \tilde{n}_{(k)} \qquad k = 1,2,...,K \tag{12}$$

Note that $P_{(k)}$ is a self-adjoint operator, $z_{(k)}$ exactly corresponds to the image by implementing the back-propagation algorithm on the data $y_{(k)}$. Now, we can regard Eq. (12) as the image processing problem.

It remains computationally intensive if solving directly Eq. (12) since it has almost the same computational complexity as solving Eq. (5). Recall that the element of matrix $P_{(k)}$, i.e., $P_{(k)}(i,j)$, describes the interaction between two resolution units, where *i* and *j* are indices of the *i*th and *j*th resolution units, respectively. As a matter of fact, $P_{(k)}$ corresponds to the system response function used in the area of radar imaging. It has been well accepted that the interaction between two distant points is ignorable compared to that between two neighbored points, which can be justified by the physical mechanism of electromagnetic wavefield. More strictly, for a voxel denoted by *i* at the location of $r_i$, the majority of energy of $P_{(k)}(j,i)$ as function of $r_j$ is concentrated around $r_i$, as illustrated in Figure 3. Inspired by this observation, we explore the idea of overlapping patches processing techniques, widely used in the area of image processing [e.g., 32], in solving Eq. (12), ignoring the affections from distant points.

Algorithm 2: the whole algorithm procedure for fast parallel-imaging *Algorithm*
___
Input : $y_{(k)}$ $(k = 1,2 \cdots K)$ and $\mathbf{A}_{(k)}$
Output : Reflectivity: $X$
___
**First step**: the whole investigation domain is divided into a series of overlapping blocks, and calculate $\mathbf{P}_{(k,b)}$ and $R_b z_{(k)}$

**Second step**: solving the sub-problems over overlapping patches by performing the first-order iterative algorithm (Algorithm I) in the parallel manner
- Do while not achieving at some stop criterion for each patch in parallel:
    For k = 1:K
  - Step 1: Compute $\nabla f(x_{(k,b)}) = d_{(k,b)}$
  - Step 2: Compute the step factor $\alpha_{(k,b)} = \dfrac{-\|d_{(k,b)}\|_2^2}{\|\mathbf{P}_{(k,b)} d_{(k,b)}\|_2^2}$
  - Step 3: Update the *n*th $x_{(k,b)}$, and $x_{(k,b)}^n = x_{(k,b)}^{n-1} + \alpha_{(k,b)} d_{(k,b)}$
    End
  - Step 4: Fuse $x_{(k,b)}$ into $\widehat{X}_{(b)}$
- End do while

**Third Step:** splicing images $\{\widehat{X}_{(b)}\}$ into the whole image $X$.
___

The flow chart of proposed imaging technique over overlapping patches is illustrated in Fig.4. The whole investigation domain is divided into a series of overlapping three-dimensional blocks (we take the name of "patch" in the area of image processing below). Note that the neighbored patches are overlapped to suppress possible artifacts around the edge of patches. For the *b*th patch, Eq. (12) becomes

$$R_b z_{(k)} \approx \mathbf{P}_{(k,b)}[R_b x_{(k)}] \qquad b=1,2,\ldots,B \qquad (13)$$

Herein, $R_b$ is the operator that extracts pixels belonging to the *b*th patch [32, 33], $\mathbf{P}_{(k,b)}$ denotes the mapping matrix associated with the *b*th patch, and *B* is the total number of patches. Correspondingly, Eq. (6) is modified as

$$min_X[\varphi_b(R_bX) + \gamma\Omega(R_bX)] \qquad b=1,2,\dots,B \qquad (14)$$

where $\varphi_b(R_bX) = \frac{1}{2}\sum_{k=1}^{K}||R_bz_{(k)} - \mathbf{P}_{(k,b)}[R_bx_{(k)}]||_2^2$, the subscript *b* highlights the overlapping processing. The solution to Eq. (14) is pursued for each patch by performing first-order iterative algorithm (Algorithm 1) in the parallel manner.

The final image is procedure by simply splicing all images of overlapping patches. For conveniences, the whole algorithm procedure is summarized in Algorithm 2. Note that $R_bz_{(k)}$ is a vector of $N_B$, and $\mathbf{P}_{(k,b)}$ is of $N_B \times N_B$. Since typically, $N_B \ll N$, like, $N_B = 10 \times 10 \times 10$ for three-dimensional problem. Consequently, the original problem Eq. (12) has been transformed into a series of smaller problems denoted by Eqs. (14), which can be solved efficiently.

## VI. Results

We hereby present various examples to demonstrate the performance of the proposed high-resolution imaging methodology in the configuration of three-dimensional MIMO microwave imaging sketched in Fig.1. Simulation data are generated by performing the commercial software of XFDTD, a full-wave Maxwell's solver. The transmitted waveform is modulated Gaussian wave with carrier frequency 2GHz and pulse width 2ns. In our numerical simulations, the whole frequency band is uniformly divided with step of 26Mhz from 0.5Ghz to 3.5Ghz. The image quality is quantitatively evaluated using the structural similarity (SSIM) widely adopted in the area of image processing [34]. All computations are performed in a small-scale server with the configuration of 32GB access memory, Intel Xeon E5-1620v2 central processing unit, and Matlab 2014 environment.

*Example 1*: Three-dimensional triple-crossed-bars

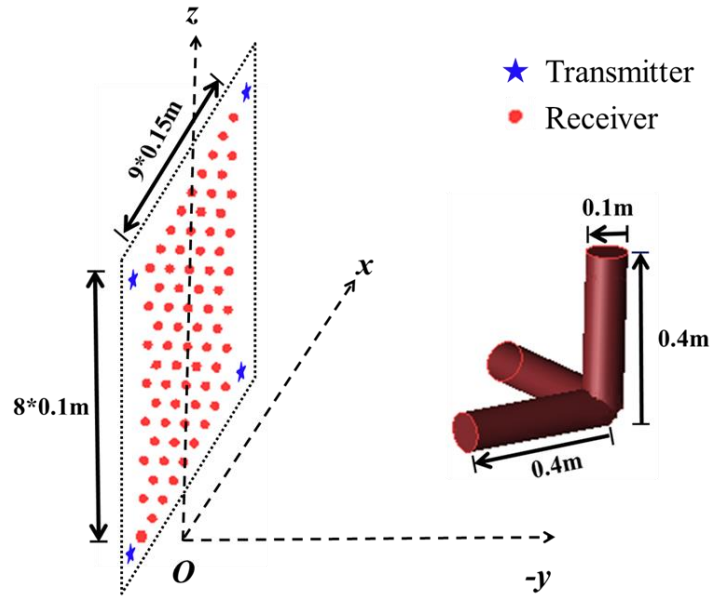

**Fig.5** The imaging configuration of example 1 is sketched and some simulation parameters are annotated.

In this example, the probed object consists of three identical conduct bars of 5cm radius and 40cm length, which are perpendicular each other and jointed together at a corner (0.047m,-0.6m, 0.25m), as sketched in Fig.5. It should be emphasized that this probed object resembles physically the corner reflector, with which important multiple scattering effects will happen instead of simple single-scattering effect. The MIMO configuration has 4 transmitters located at (-0.61 m, 0 m,0 m), (0.61 m, 0 m,0 m), (-0.61 m, 0 m,0.7 m), and (0.61 m, 0 m,0.7 m) as marked in blue stars, and a uniform array of $9 \times 8$ receivers with space of 0.15 m and 0.1 m, respectively. Since the reduced currents within the object vary importantly with different transmitters, we assume that one component of generalized reflectivity, i.e., $x_{(k)}$, corresponds to one transmitter, consequently, *K*=4. Additionally, the investigation domain *V,* which is centered at (0 m, -0.6 m, 0.5 m) with size of 2 m by 0.8 m by 1 m, is divided uniformly into 40 by 16 by 20 sub-grids, and each subgrid has the size of 0.05 m by 0.05 m by 0.05 m. Reconstruction results calculated using different methods are compared in Fig.6, where corresponding SSIM values for Figs. 6 (a)-(f) are 0.4994, 0.5102, 0.5741, 0.7826, 0.7836 and 0.7762, respectively. This reflects that the proposed GR-based method is better than other methods in terms of the quality of obtained images.

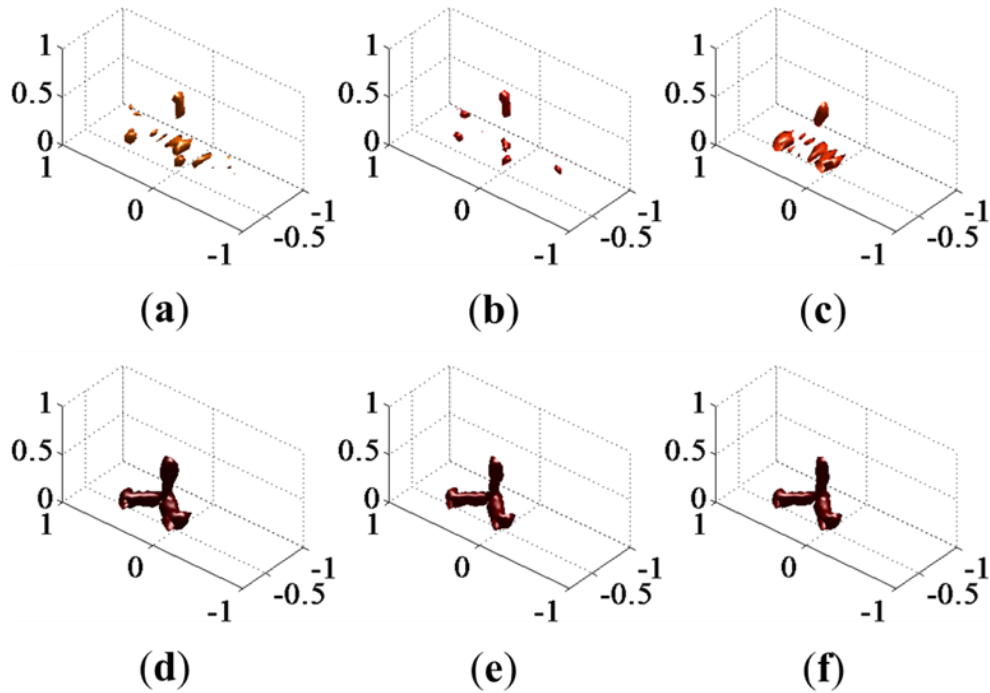

**Figure 6**. The reconstruction results of example 1. Figs. 6 (a) and (b) present the results produced by the time-domain filtered back-propagation algorithm and time-domain first-order iterative algorithm, respectively. Figure 6 (c) shows the result by using the standard Born-based model (equivalently, K=1). Figures 6 (d)-(f) report the reconstructed results based on the generalized reflectivity model in combination with (1,2)-, (1,1)-, and (1, ∝)- mixed norms of Eq. (5), respectively, where the first-order iterative algorithm ( Algorithm 1) is used. In these figures, all results are normalized by their own maximum. (Unit: m)

From this set of results, several conclusions can be immediately drawn, in particular:

(a)Overall, the proposed GR-based method has the image quality superior to other methods, which does make sense since the GR model accounts for more realistic interaction of wavefield with the probed object, as emphasized above. By using the proposed imaging methodology, three arms of probed object with clear pattern can be clearly reconstructed. Additionally, the convergence rates for different choices of (p,q)-mixed-norm are almost the same, as demonstrated in Fig. 7.

(b) In terms of image quality, the back-propagation algorithm produce the worst image of Fig. 6 amongst all results, where some artifacts exist and the shape of three-arm object has been seriously distorted.

(c) Although the Born-based imaging can produce the result (Fig. 6c) slightly better than that by the back-propagation algorithm (Fig.6a), careful readers can notice that the reconstructed image consists of some discrete points, and the entire smooth pattern or structure of probed object has been broken, which possibly comes from ignoring the multiple scattering effects.

(d) From the viewpoint of computational complexity, the back-propagation algorithm is the lowest since it only involves the operation of matrix-vector multiplication once, specifically, it only took 0.68 seconds in this example. However, the iterative methods, either the Born-based model or the generalized reflectivity model, have relatively higher computational complexity due to the implementation of a large amount of matrix-vector multiplications. For this example, their computational time is around 2 minutes. Additionally, it is noted from Fig. 7 that the convergence rate of proposed GR-based method is slightly worse than that of Born-based method, since the former has more unknowns than the latter, as mentioned previously.

Above results demonstrate that the GR-based imaging is superior to the conventional Born-based method in terms of image quality, since it accounts for more complicated interaction between the wavefields and objects. Furthermore, the GR model can handle the object with anisotropic scattering properties, in contrast to Born model limited to isotropic objects. To see it clearly, four reconstructed image in four different view angles based on the generalized reflectivity model are shown in following Fig.8. It is noted that four reconstructed results are different and worse than results in the bottom line of Fig.6, which verifies our hypothesis.

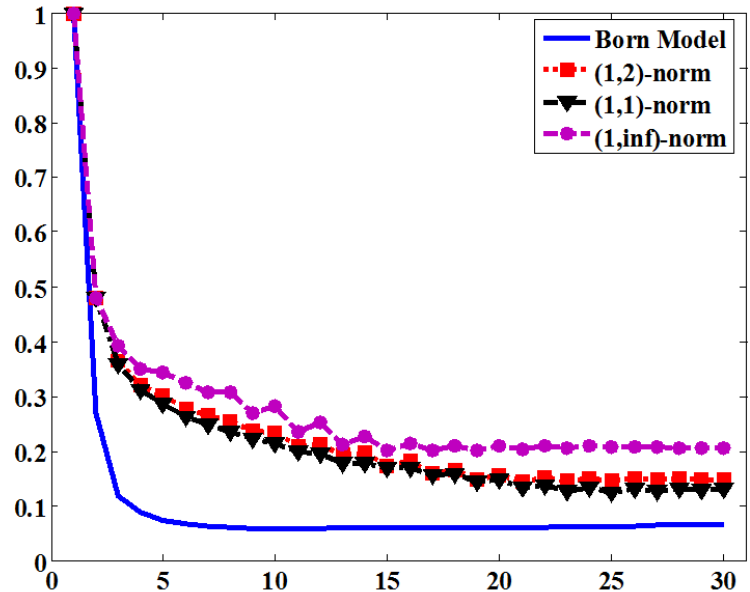

**Fig.7** The convergence curves of Born model (blue) and the generalized reflectivity model in combination with (1,2)-, (1,1)-, and (1, ∝) mixed norms of Eq.(5).

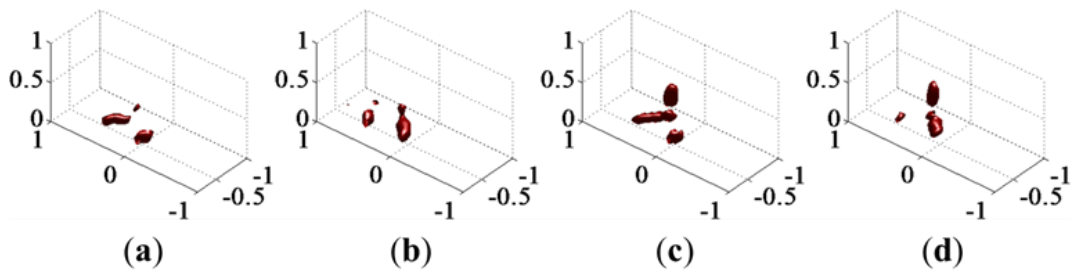

**Figure.8** The reconstructions from four different view angles (transmitters). (Unit: m)

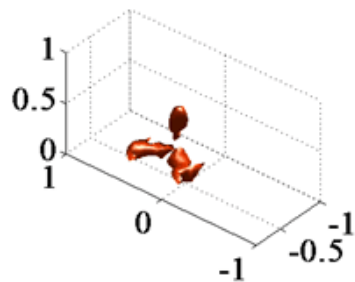

**Figure.9** The reconstructions from four different view angles (transmitters) and three overlaping sub-bands. Other simulation parameters are the same as those used in Fig. 6. (Unit: m)

Finally, we would like to say that, in principle, the imaging quality should be more accurate when the frequency sub-band is taken into account. However, the implementation of frequency sub-bands will introduce more additional unknowns, worsen the ill-posedness degree of Eq. (5). For this reason, the imaging quality will be degenerated, as illustrated in Fig. 9. So the imaging accuracy and computational complexity should be trade-off well.

***Example 2***: *Three-dimensional cartoon human body model*

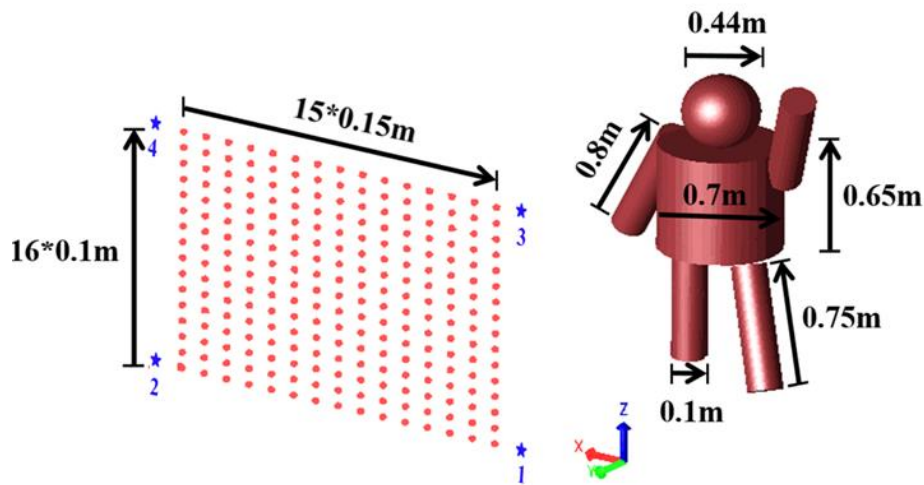

**Fig.10** The imaging configuration of example 2 is sketched. (Unit: m)

Here we examine the performance of proposed methodology using a more visually complicated scenario, a three-dimensional cartoon human body, as illustrated in Fig. 10. The simulated human body consists of a sphere with radius of 0.44 m as head, a cylinder with radius of 0.35 m and length of 0.65 m as main chest, two cylinders of 0.05 m radius and 0.8 m length as arms, and two cylinders of 0.05 m radius and 0.75 m lengths as legs. The overall height of human body is 1.8 m. Additionally, the whole body has relatively dielectric parameter of 50. The

MIMO configuration has four transmitters marked in blue and a uniform array of 15 by 16 receivers with space of 0.15 m along x direction and 0.1m along z direction. The investigation domain V, which is centered at (0 m, -2 m, 0.85 m) with size of 4 m by 3 m by 2.5 m, is divided uniformly into 40 by 30 by 25 sub-grids, and each has the size of 0.1 m by 0.1 m by 0.1 m. Other simulation parameters are the same as those used in Example 1. The results obtained by different algorithms are compared in Fig. 11.

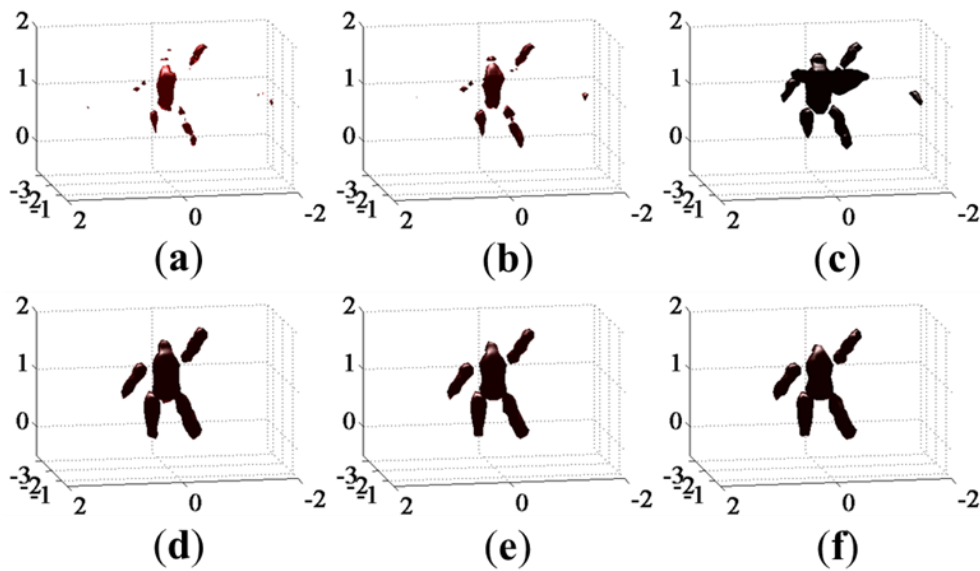

**Fig.11** The results reconstructed using the same algorithms as example 2. Figure 12 (a)-(b) provide the results produced by performing the time-domain filtered back-propagation algorithm and the time-domain first-order iterative algorithm, respectively, while Figure 12 (c) shows the result by using the standard Born-based model (equivalently, K=1). Figures 12 (d)-(f) report the reconstructed results based on the generalized reflectivity model in combination with (1,2)-, (1,1)-, and (1, $\infty$)- mixed norms of Eq. (5), respectively. In addition, the SSIM values for Fig.12 (a)-(f) are 0.5749, 0.6715, 0.7138, 0.7325, 0.7346 and 0.7428 respectively. (Unit: m)

From this set of results, the conclusions drawn previously can be verified, i.e., (a) the imaging methodology based on the generalized reflectivity model is superior to both the Born model and the back-propagation algorithm in terms of image quality. By using the proposed imaging methodology, the shape of cartoon human body can be easily identified. (b) The back-propagation algorithm produce the worst image amongst all results, where some artifacts exist and the shape of cartoon human body cannot be identified. (c)Although the Born model can produce the result better than that by back-propagation algorithm, some artifacts remains being occurrence around the main chest body, which possibly comes from multiple scattering effects. The back-propagation algorithm has lower computational complexity, it took about 13 minutes for this example. However, the imaging methodologies using both the generalized reflectivity model and the conventional Born model have relatively higher computational complexity and require higher hardware configuration due to the iterative implementation.

***Example 3****: Fast imaging of three-dimensional cartoon human body*

Above results demonstrate that compared to the back-propagation algorithm, Algorithm 1 has heavy computational cost since it involves all measurements and unknowns at each iteration. The computational complexity will increase drastically as the growth of unknowns, therefore it is computationally expensive for treating large-scale problems. To overcome this drawback, we developed a fast imaging algorithm by transforming the imaging problem into a physics-driven image processing problem in combination with the overlapping processing technique in section II. Here, we examine its performance by imaging the three-dimensional cartoon human body investigated in Example 2. The whole discrete investigation domain is uniformly partitioned into a series of patches with half overlap. Figures 12 reports the images reconstructed using Algorithm 2 for different number of overlapping patches, i.e., 1 (no patch processing), 3, 9, 27, 63, and 147, respectively. In addition, the computation time and SSIM value for different choices of number of patches are compared in table 2. Moreover, figure 13 illustrates some results from 16 patches out of 27 patches in Fig. 12(d). From this set of results, it can be concluded that when the number of partition patches is increased (i.e., the size of patch is decreased), the computation time will be decreased at the cost of sacrificing image quality. For instance, if 147 patches are used, the computational time can be will be significantly reduced, however, the quality resulted image is relatively as shown in Fig. 12(f). In particular, some artifacts occur around cartoon body, and its associated SSIM is relatively low. On the contrary, if the smaller number of patches is used, for example, 3 parches are used, the image quality will be remarkably improved at the cost of sacrificing computation time, as illustrated by Fig. 12(b) and its associated SSIM being 0.7648. In this sense, there is trade-off between the imaging speed and accuracy. Nonetheless, the block-wise GR-based imaging technique could be applicable to large-scale microwave imaging problem.

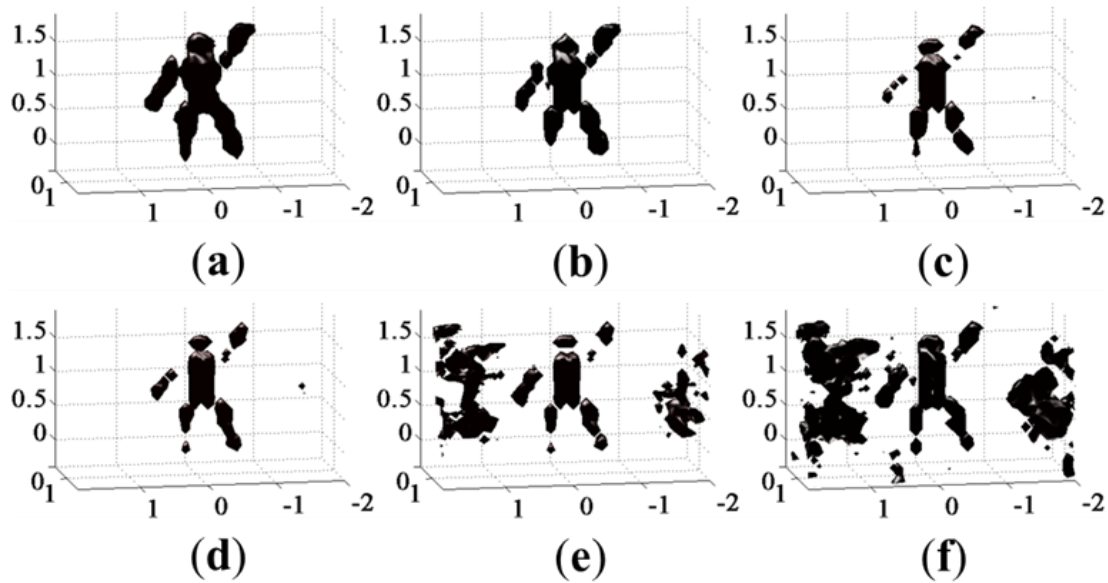

**Fig.12** The images (a)-(f) reconstructed using Algorithm 2 for different number of sub-blocks, i.e., (a)1 patch,(b) 3 patches, (c)9 patches, (d)27 patches, (e) 63 patches, and (f)147 patches. (Unit: m)

Table 1. Comparison of computation time and SSIM for different overlapping patches in Fig. 13.

| Number of patches | Computation time(s) | SSIM |
| --- | --- | --- |
| 1 | 2193 | 0.7782 |
| 3 | 77 | 0.7648 |
| 9 | 8 | 0.7563 |
| 27 | 2 | 0.7502 |
| 63 | 0.5 | 0.6680 |
| 147 | 0.1 | 0.6630 |

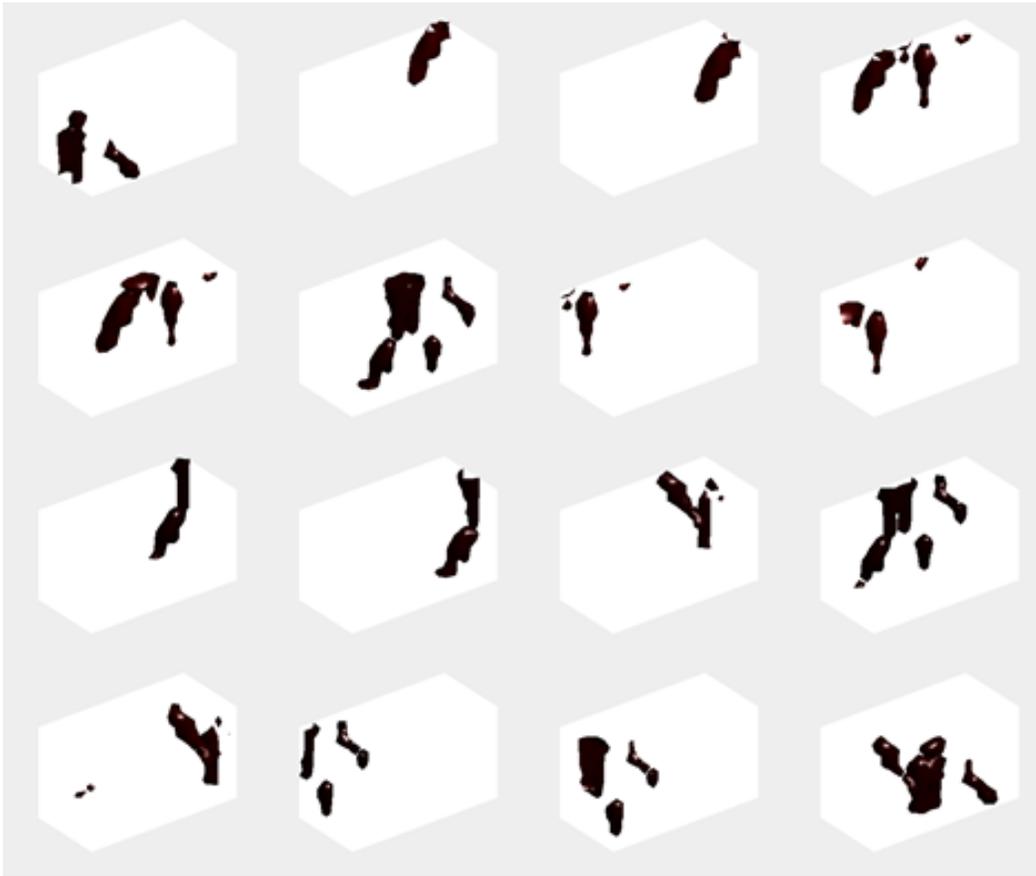

**Fig 13.** 16 patch results out of 27 corresponding to Fig. 12(d).

## V. Conclusions

Generally, microwave imaging is an important tool in detecting and identifying objects in nondestructive manner and it has been widely applied to many scenes in civil or military. However, the low quality image and high computation cost set an impenetrable barrier for its further development. We introduce the generalized reflectivity as a function of operational frequencies and viewing angle. In this way, this model extends the conventional Born-assumption into the one taking more complicated interactions between the electromagnetic wavefields and the investigated objects into account. In this sense, this model could have the potential of producing high-quality image. Compared to Born-based imaging techniques, the use of generalized reflectivity will introduce a large amount of additional unknowns, making the undergoing imaging problem more ill-posed. To address this issue, the imaging problem has been formulated in the framework of sparsity-promoted optimization. The major difficulty faced by the resulting problem is very expensive computational cost. To overcome this issue, we turn the imaging problem into the problem of physics-driven image processing. Moreover, the image processing is performed over overlapping patches, which can be very efficiently performed in the parallel or distributed manner. Therefore, it can be expected that the proposed imaging methodology is applicable to large-scale high-resolution imaging problems. Some selected simulation results are provided to demonstrate the state-of-art performance of proposed methodology.